\documentclass{article}

\pdfoutput=1
\usepackage[preprint]{neurips_2024}

\usepackage[utf8]{inputenc} % allow utf-8 input
\usepackage[T1]{fontenc}    % use 8-bit T1 fonts
\usepackage{url}            % simple URL typesetting
\usepackage{booktabs}       % professional-quality tables
\usepackage{amsfonts}       % blackboard math symbols
\usepackage{nicefrac}       % compact symbols for 1/2, etc.
\usepackage{microtype}      % microtypography
\usepackage{xcolor}         % colors

\usepackage{times}
\usepackage{soul}
\usepackage{url}
\usepackage{amsmath}
\usepackage{amsthm}
\usepackage{booktabs}
\usepackage{algorithm}
\usepackage{algorithmic}
\usepackage{multirow}

\usepackage{amsmath,amssymb,amsfonts}
\usepackage{stmaryrd}
\usepackage{hhline}
\usepackage{booktabs, caption, tabularx, makecell}
\usepackage{xcolor}
\usepackage{graphicx} % DO NOT CHANGE THIS
\usepackage{caption} % DO NOT CHANGE THIS AND DO NOT ADD ANY OPTIONS TO IT
\usepackage{tcolorbox}
\tcbuselibrary{theorems}

\newtcbtheorem{mytheo}{Proposition}%
{colback=white,colframe=gray,fonttitle=\bfseries}{th}
\usepackage{algorithm}

\usepackage{algorithmic}

\usepackage{booktabs, caption, tabularx, makecell}

\usepackage{comment}
\usepackage{xcolor}

\newcommand{\re}[1]{{\color{black}#1}}
\newcommand{\man}[1]{\textcolor{red}{\textit{[\textbf{Man}: #1]}}}

\usepackage{subcaption}

\usepackage{bibentry}

\newtheorem{prop}{Proposition}

\usepackage{newfloat}
\usepackage{listings}

\usepackage{wrapfig}

\title{Enabling Tensor Decomposition for Time-Series Classification via \re{A Simple Pseudo-Laplacian Contrast}}
\author{%
  Man Li \\
  Southwestern University of Finance and Economics\\
  Georgia Institute of Technology\\
  Chengdu, China and Atlanta, GA, U.S. \\
  \texttt{mlicn@connect.ust.hk} \\
  \And
  Ziyue Li \\
  University of Cologne \\
  Cologne, Germany \\
  \texttt{zlibn@wiso.uni-koeln.de} \\
  \AND
  Lijun Sun \\
  McGill University \\
  Montreal, Canada \\
  \texttt{lijun.sun@mcgill.ca} \\
  \And
  Fugee Tsung \\
  The Hong Kong University of Science and Technology \\
  Hong Kong \\
  \texttt{season@ust.hk} \\
}

\begin{document}

\maketitle

\begin{abstract}
Tensor decomposition has emerged as a prominent technique to learn low-dimensional representation under the supervision of reconstruction error, primarily benefiting data inference tasks like completion and imputation, but not classification task. 
We argue that the non-uniqueness and rotation invariance of tensor decomposition allow us to identify the directions with largest class-variability and simple graph Laplacian can effectively achieve this objective.
Therefore we propose a novel Pseudo Laplacian Contrast (PLC) tensor decomposition framework, which integrates the data augmentation and cross-view Laplacian to enable the extraction of class-aware representations while effectively capturing the intrinsic low-rank structure within reconstruction constraint.
An unsupervised alternative optimization algorithm is further developed to iteratively estimate the pseudo graph and minimize the loss using Alternating Least Square (ALS).
Extensive experimental results on various datasets demonstrate the effectiveness of our approach.

\end{abstract}

\section{Introduction}

\begin{figure*}[t]
\centering
\includegraphics[width=1\textwidth]{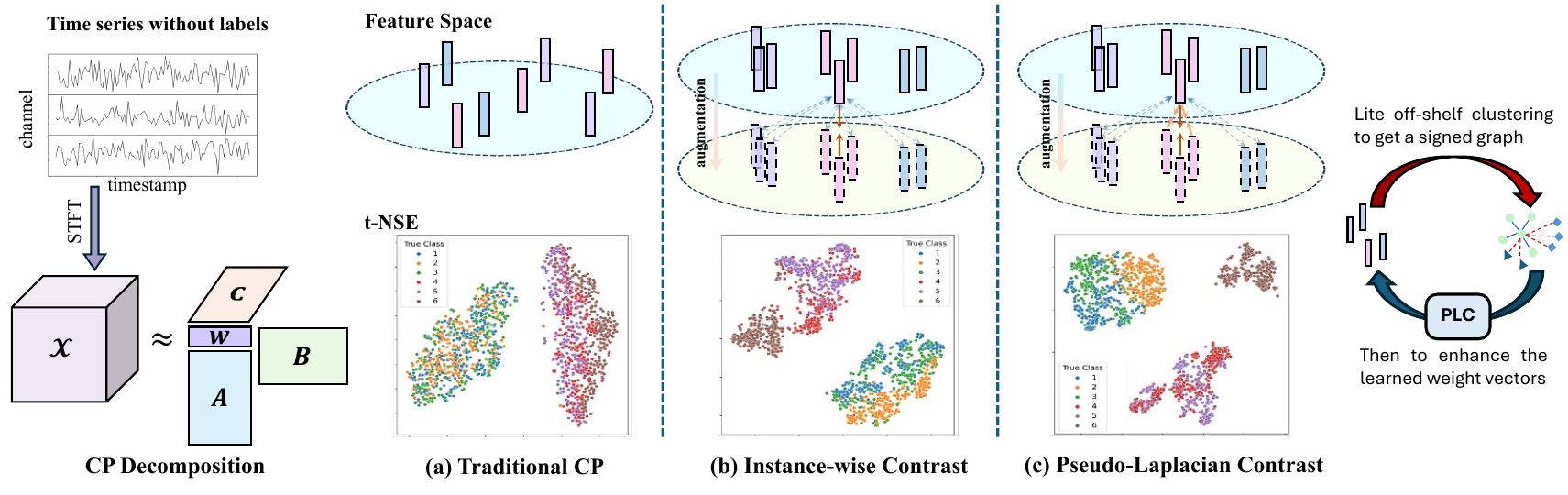}
\caption{The class information in the feature space: (a) the features extracted by CP decomposition are messed together; (b) instance-wise contrast (e.g., ATD \citep{y:22} and InfoNCE \citep{oord2018representation}) treating the data from the same instances as a positive pair only presents obscure class patterns; (c) cross-view Laplacian iteratively learn pseudo graph and use cross-view Laplacian to enhance the class-distinctiveness. ( $\dashleftarrow\dashrightarrow$: pull far away, $\rightarrow\leftarrow$: push closer with darker color indicating more impact).} 
% \ziyue{(1) no need to draw the different class's negative samples, maybe only highlight the purple class?; (2) highlight the label is unknown, so we need to learn a pseudo graph to have the light-orange links; (3) add fig (d): the pseudo graph and feature are cyclically improving each other}}
\label{fig:motivation}
\vspace{-15pt}
\end{figure*}

\textbf{Tensor decomposition is known for its suboptimal performance in classification tasks} \citep{y:22}:  Tensor decomposition, originally as a dimension-reduction method, although has been successfully applied to multi-dimensional time-series data, such as in neuroscience \citep{cong2015tensor}, transportation \citep{chen2019bayesian} and economics \citep{wang2022high}, is mainly fitted for regression-like tasks: prediction \citep{yan2024sparse, li2020tensor,li2020long} and imputation \citep{hu2024low, chen2019bayesian}. The reason is stemmed from the way how the decomposition is learned: take the CANDECOMP/PARAFAC (CP) tensor decomposition \citep{k:09} on a three-dimensional tensor $\mathcal{X}$ as an example, $\mathcal{X}$ is decomposed into three mode matrices along each dimension, $\mathbf{A,B,C}$, with a weight vector $\mathbf{w}$; however, \re{as shown in Fig. \ref{fig:motivation}} (a), $\mathbf{w,A,B,C}$ are only learned by minimizing the \textit{reconstruction loss} $\| \mathcal{X} - \llbracket  \mathbf{w,A,B,C} \rrbracket \|_2^2$ (with details later). This creates two problems of the decomposition results: nonuniqueness and rotation invariance \citep{k:09}, e.g., the decomposed $\mathbf{w}$ can be scaled, transformed, or rotated in each decomposition, even under the same reconstruction loss. This won't be a problem for regression-like tasks, but it will significantly affect the classification task since most of the tensor-based classification paradigms uses the decomposed $\mathbf{w}$ as the feature vector of each sample. As a result, shown in Fig. \ref{fig:motivation}(a), the features are learned with all class patterns messily mixed.

\re{Instance-wise contrastive learning \citep{oord2018representation, zhu2022tico} is popular for  unsupervised task to enhance \textit{alignment} and ATD \citep{y:22} first attempt to integrate such contrastive loss with tensor decomposition for class-aware feature learning. Despite its improved performance, treating $\mathbf{w}$ and its direct augmented sample $\widetilde{\mathbf{w}}$ as positive and others as negative regardless of the latent class, as shown in Fig \ref{fig:motivation}(b), may harm the performance (\textbf{-2.39\%} worse than ours).
Existing literature have emphasized the importance of negative sampling for contrastive learning \citep{arora2019theoretical, robinson2020contrastive}. 
Clustering based models have been proposed to involve pseudo labeling \citep{caron2018deep} or prototypes \citep{li2020prototypical} to capture semantic similarity, however the pseudo class information is commonly used on single intra-view.} 
% where only $\mathbf{w}$ and its direct augmented sample $\widetilde{\mathbf{w}}$ are positive sample: this is reasonable since class labels are unknown when training the tensor decomposition, then only a sample and its augmented one can be 100\% guaranteed as positive. However, such an instance-wise alignment may even harm the pattern learning: (1) as shown in Fig \ref{fig:motivation}(b), it will treat other samples from the same class as negative samples, which harms the performance (\textbf{-\#\%} worse than ours); (2) we prove that such a instance-contrast loss is a lower bounded by ours, showing that its intra-class is less compact and inter-class is less separate as ours, as we can see by comparing Fig. \ref{fig:motivation}(c) with (b) \man{ %(2) its class-aware distinctiveness in the feature space will be obscured without considering both the inter- and intra-class distances;
% (3) ignoring of the diversity of samples within the same class may lead to over-fitting issue, thereby reducing generalizability.} e.g., due to a limited knowledge of a class, it needs more \textbf{$\times$\# times} more samples for supervised training of the classifier to achieve similar result as ours
% %\ziyue{being prone to overfitting and lacking generalizability? since the positive samples ignore the diversity of samples from a same class}; (3) anymore? (same as the disadvantage of instance SSL vs block ssl)

% \re{nonuniqueness and rotation invariance (toy experiment: repeat the CP decomposition for N times and plot their t-sne to see whether some have better class information) -> class-alignment}

To address the weakness of \re{existing methods}, \re{we propose a simple yet effective solution from the perspective of Laplacian penalty}: \textbf{can we construct a pseudo graph indicating which samples are from the same class based on learned features, and then use the graph to improve the feature in return}? Specifically, apart from the reconstruction loss, such a pseudo graph is introduced as auxiliary information \re{by the well-designed cross-view graph Laplacian over raw data and class-preserved augmented data to simultaneously mitigate class-irrelevant noise and emphasize the class-distinctive information.}
% guide the graph contrast, and augmentation-based instance contrast is also included to mitigate class-irrelevant noise. 
Then, we jointly optimize the contrastive objectives and tensor decomposition reconstruction objective using Alternative Least Square (ALS).

\re{

As a result, via such an iterative updating between pseudo graph and feature, similar to the EM algorithm: (1) not only our features are improved to preserve more class pattern (with around \textbf{3\%} higher classification accuracy than CP and clearer separation in the embedding space in Fig. \ref{fig:vis}(b)), (2) the learned pseudo graph is also close to the ground truth (in Fig. \ref{fig:vis}(a)). This is also proved by the small margin compared with using the ground truth label for guidance, instead of pseudo graph. (3) We show that the pseudo graph contrast is similar to the classical InfoNCE framework \citep{oord2018representation}, but with a different positive sample definition according to our leaned pseudo graph, and also enjoys the block-wise contrastive property of being more compact and preserving more class information. (4) We also show a great insight that the tensor decomposition is a two-edge sword as shown in Fig. \ref{fig:overview}(b); as mentioned before, the decomposition result is not identifiable and rotation-invariant: firstly, due to the non-uniqueness, the reconstruction loss provides a bunch of non-identifiable decomposition results with a small $\epsilon$ region; then, thanks to the rotation-invariance, our pseudo graph contrast is actually ``rotating'' the decomposition result to an angel that maps the class pattern to the most while maintaining the same reconstruction loss. 

}

\section{Preliminaries}

\paragraph{\textbf{Notations.}} 
Let $\mathcal{X}\in \mathbb{R}^{I_1 \times \cdots \times I_N}$ denote the $N$-order tensor where $x_{i_1,\dots i_N}$ is the $(i_1,\dots i_N)$-th element and $\mathbf{X}_{(k)}\in \mathbb{R}^{I_k \times \prod_{j\neq k}I_j}$ is the mode-$k$ unfolding.
The $k$-th row and column of the matrix $\mathbf{X}$ are represented as $\mathbf{x}^{(k)}$ and $\mathbf{x}_k$ respectively.
The vector and scalar are denoted by lowercase bold and plain fonts, e.g., $\mathbf{x}, x$, respectively.
We use $\|\cdot\|_F$ to denote the Frobenius norm, and $\langle\cdot,\cdot\rangle$, $\circ$, $\llbracket \cdot \rrbracket$, and $\odot$ for the inner product, outer product, Kruskal product and Khatri-Rao product respectively.

\paragraph{\textbf{Tensor modeling.}}
High-dimensional time series data analysis relies on the assumption of an inherent low-rank structure for efficient computation.
In this context, we assume that $N$ preprocessed time series samples 
$\mathcal{X}^{(n)} \in \mathbb{R}^{I_1 \times \cdots \times I_K}$ 
are stored in $\mathcal{X}=\left[\mathcal{X}^{(1)}, \ldots, \mathcal{X}^{(N)}\right] \in \mathbb{R}^{N \times I_1 \times \cdots \times I_K}$, where $I_k$ can generically mean multi-dimensions such as timestamp and other features. 
% For example, we define $I,J$ and $K$ represent $\textit{channel}$, $ \textit{frequency}$, and $\textit{timestamp}$ respectively.
% For tensor data, it is common to assume the low-rank structure, and 

\textbf{CANDECOMP/PARAFAC (CP) decomposition}:
Following one of the most popular techniques, CP decomposition \cite{k:09}, given the $n$-th sample $\mathcal{X}^{(n)}$, its low-rank structure can be represented as the approximation of a weighted sum of rank-one tensor, such that,
\begin{equation} \label{eq:cp}
    \mathcal{X}^{(n)} = \sum_{r=1}^R w_r^{(n)} \mathbf{u}_r^{(1)} \circ \mathbf{u}_r^{(2)} \circ \cdots \circ \mathbf{u}_r^{(K)} = [\![\mathbf{w}^{(n)}; \mathbf{U}^{(1)}, \mathbf{U}^{(2)}, \cdots, \mathbf{U}^{(K)}]\!],
\end{equation}
where $R$ is the rank of tensor, $\mathbf{U}^{(k)} = [\mathbf{u}_1^{(k)},...,\mathbf{u}_R^{(k)}] \in\mathbb{R}^{I_k\times R}$ is the decomposed factor matrices along the $k$-th mode, $\mathbf{w}^{(n)}\in\mathbb{R}^{1\times R}$ is the feature vector.
CP decomposition result is non-uniqueness but rotational invariant, and we denote the feasible feature space given a set of factor matrices as $\mathcal{E}$.

% \textbf{$\epsilon$ - Approximation}:
% % Given the basis $ (\mathbf{U}^{(1)}, \mathbf{U}^{(2)}, \mathbf{U}^{(3)}) $, 
% Instead of learning the best approximation, we consider the border $\epsilon$ - approximation region defined as:
% \begin{equation} \label{eq:border}
%     \mathcal{E}(\epsilon) = \{\mathbf{w}^{(n)}:   \|\mathcal{X}^{(n)} - [\![\mathbf{w}^{(n)}; \mathbf{U}^{(1)}, \mathbf{U}^{(2)}, ..., \mathbf{U}^{(N)}]\!]\|_F < \epsilon, \forall n \ \ | (\mathbf{U}^{(1)}, \mathbf{U}^{(2)}, \mathbf{U}^{(3)})\}.
% \end{equation}

% \subsection{Signed graph}
% Let $\mathcal{G} = (V, E)$ be a signed graph and $\mathbf{S}$ is the corresponding symmetric weight matrix with arbitrary elements and zero diagonal entries \cite{gallier2016spectral}. The signed degree matrix and normalized graph Laplacian matrix can then be defined as:
% \begin{equation}
%      \mathbf{D}  = \operatorname{diag}(d_1,\cdots,d_N) \text{ with } d_i = \sum_{j} |S_{ij}|,\ \ \ \ \ 
%      \mathbf{L}  = \mathbf{I} - \mathbf{D^{-1/2}GD^{-1/2}} \in \mathbb{S}_+,
% \end{equation}
% where $\mathbf{L}$ is symmetric positive semi-definite. 
% % The typical contrastive loss for graph is given by:
% % \begin{equation}
% %     \mathcal{L} = \mathrm{E}_{v}[ \mathrm{E}_{u\sim p_d(u|v)}sim(u,v) + \phi \mathrm{E}_{u' \sim p_n(u' |v)}sim(u' ,v)],
% % \end{equation}
% % where $u$ and $u'$ are the positive and negative sample nodes for the given anchor node $v$ \cite{zhu2021contrastive}.

\paragraph{\textbf{Problem formulation.}}
Given one time series sample $\mathcal{X}^{(n)}$, it will first go through our proposed unsupervised representation extractor PLC to obtain the feature vector and then a supervised classifier $g(\cdot)$ for class inference, i.e., $p | \mathcal{X}^{(n)} \sim g(\text{PLC}(\mathcal{X}^{(n)}))$. 

\section{Methodology} \label{sec:method}

\begin{figure*}[t]
\centering
\includegraphics[width=\textwidth]{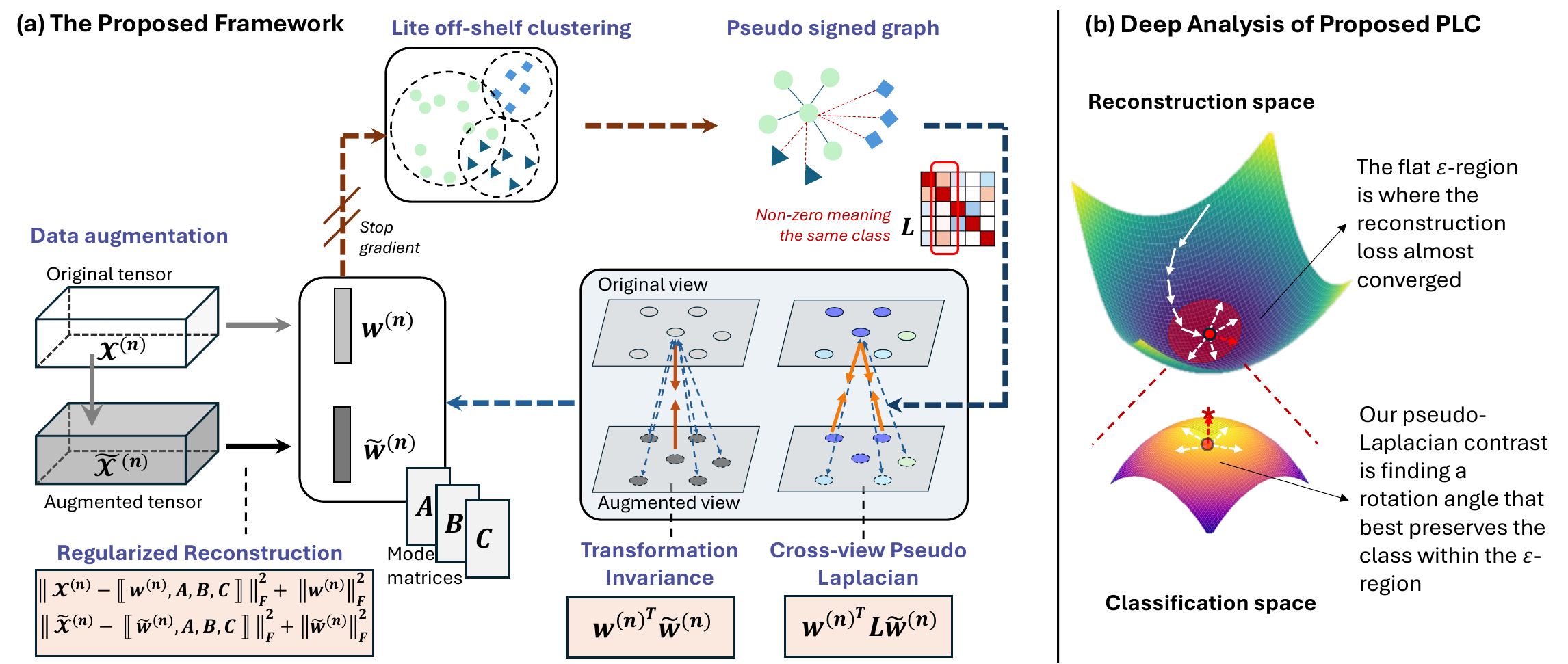}
\caption{(a) Original $\mathcal{X}$ is \textbf{augmented} to $\widetilde{\mathcal{X}}$; both are \textbf{decomposed} into shared matrices $\mathbf{A,B,C}$ and their own feature vectors $\mathbf{w}, \widetilde{\mathbf{w}}$. An \textbf{off-shelf clustering} groups the features and gets a pseudo-signed graph indicating class relations. The \textbf{pseudo-Laplacian contrast} pushes the cross-view features from the same pseudo-class (colored) together and different classes away; The \textbf{transformation invariance} (need no pseudo-class, grey) only pushes the same sample together and different samples away; (b) When reconstruction converges in a small region, decomposition results are non-identifiable; thanks to the rotation-invariance, our PLC finds a best rotating angle to best present the class information.}
%Loss function has three terms: (i) pseudo-graph contrastive loss, (ii) instance contrastive loss, and (iii) reconstruction loss with regularization on $\mathbf{w}$.  Here, $\mathcal{X}^{(n)}$ represents the time series data after Short-Time Fourier Transforms, while $\widetilde{\mathcal{X}}^{(n)}$ is its augmented counterpart. Additionally, $\mathbf{w}^{(n)}$ and $\widetilde{\mathbf{w}}^{(n)}$ are their corresponding feature matrices. $\mathbf{A}, \mathbf{B}$, and $\mathbf{C}$ are the factorization matrices.
% \ziyue{(1)change to 'Transformation Invariance' and 'Cross-view Pseudo Laplacian'; (2) the L matrix shouldn't have diagonal}}
\label{fig:overview}
\vspace{-16pt}
\end{figure*}

Traditional tensor decomposition methods focus on projecting data into low-dimensional spaces by fitting individual data samples with statistical reconstruction error measures, such as the Frobenius norm. \re{Due to the non-uniqueness of the decomposition, there will be a region of solutions satisfying the same lowest reconstruction loss}. However, for time series data with inherent class information, these decomposition methods fail to address the class-aware low-dimensional space projection.
We \re{propose novel pseudo-graph cross-view Laplacian approach that helps find a subspace that rotates to best preserve the class-related information while ensuring reconstruction accuracy}. The proposed Pseudo-graph Laplacian Contrast (PLC) framework is shown in Fig. \ref{fig:overview}.

\subsection{PLC framework}

\re{
The proposed PLC mainly has a cyclical learning as shown in Fig. \ref{fig:motivation}(c) and Fig. \ref{fig:overview}(a): (1) To learn a pseudo signed graph using the features $\mathcal{G} | w$, and (2) To enhance the learned features by PLC $w | \mathcal{G}$.
}

% \subsubsection{To Learn A Pseudo Signed Graph}

\re{

% The proposed PCL \ziyue{Or PGC?} framework mainly has a cyclical learning as shown in Fig. \ref{} \ziyue{the cycle subplot in introduction} and Fig. \ref{fig:overview}(a): (1) To learn a pseudo signed graph using the features (arrows in red) $\mathcal{G} | w$, and (2) To enhance the learned features by PGC $w | \mathcal{G}$.

\subsubsection{To learn a pseudo signed graph} 

}
\re{
Intuitively, data samples from the same class should have similar feature vectors $\mathbf{w}$, while those from different classes should have dissimilar ones in the feature space, which just matches the goal of graph Laplacian.
Therefore, we first construct a signed graph $\mathcal{G} = \{V, E\}$ based on the latent class information, with $|V| = N$ nodes corresponding to the samples and adjacent matrix $\mathbf{S}$ defined as:
\begin{equation} \label{eq:S}
    S_{ij} =\left\{
\begin{aligned}
1 & , & \text{if } \mathcal{X}^{(i)} \text{ and } \mathcal{X}^{(j)} \text{ are from the same class and }i\neq j, \\
-1 & , & \text{if } \mathcal{X}^{(i)} \text{ and } \mathcal{X}^{(j)} \text{ are from the different class and } i\neq j\\
0 & , & \text{otherwise}
\end{aligned}
\right.
\end{equation}
}

\re{
\textbf{Pseudo labeling by clustering.} To avoid over-design, we choose a lite and off-the-shelf clustering method to estimate the class information. 
%To better construct data pair samplers, we aim to estimate the class information via clustering. 
Specifically, we perform clustering techniques, e.g., $k$-means, on the feature vectors $\mathbf{w}$ to obtain pseudo labels. and then construct the pseudo-signed graph as defined by Eq. \ref{eq:S}.
Notably, the quality of the pseudo labeling theoretically highly affects the performance of the following graph Laplacian module. However, we claim that tensor decomposition reconstruction objective provides a reliable searching region for feature learning, which guarantees the consistency and stability of the pseudo-graph Laplacian module for class-aware feature extraction. In our experiment, we show that our learned pseudo graph always converge to the ground true graph under varying random seeds.
% set four different random seeds with different decomposition results, they all converge to a same signed graph, blabla} \ziyue{maybe we need to show that although different random seeds of decomposition, our learned pseudo graph is stable and always close to the ground truth} %The entire unsupervised pseudo-labeling process is conducted under the constraints of reconstruction loss, which can guarantee the performance and 
More details will be discussed in subsection \ref{sec:opt}.

}

\re{

\subsubsection{To enhance the feature via pseudo-graph Laplacian}
}

\re{
\textbf{Traditional Graph Laplacian.} Given the learned pseudo graph $\mathcal{G}$ from the last step, we have the intuitive graph Laplacian \cite{gallier2016spectral} on normalized features formulated as follows, known as Laplacian regularization: it aims to maximize the within-class similarity and minimize the between-class similarity in feature space. Since two subjects are from the same original data, we name it as same-view Laplacian loss.
\begin{equation}  \label{eq:laplacian}
    \ell_{\textit{same-view}}  = \operatorname{Tr}(\mathbf{W}^T\mathbf{\Lambda}(\mathbf{W})\mathbf{L}\mathbf{\Lambda} (\mathbf{W})\mathbf{W}) \ \propto\  
     - \sum_{i,j=1}^N S_{ij}\cdot \langle \frac{\mathbf{w}^{(i)}}{\left\|\mathbf{w}^{(i)}\right\|_{2}}, \frac{\mathbf{w}^{(j)}}{\left\|\mathbf{w}^{(j)}\right\|_{2}} \rangle ,
\end{equation}

where $\mathbf{\Lambda}(\mathbf{W})=\operatorname{diag}\left(1 /\left\|\mathbf{w}^{(1)}\right\|_{2}, \ldots, 1 /\left\|\mathbf{w}^{(N)}\right\|_{2}\right)$ is the row-wise scaling matrix; $\mathbf{L}$ is the degree normalized signed Laplacian matrix defined as:
\begin{equation} 
     \mathbf{D}  = \operatorname{diag}(d_1,\cdots,d_N) \text{ with } d_i = \sum_{j} |S_{ij}|,\ \ \ \ \ 
     \mathbf{L}  = \mathbf{I} - \mathbf{D^{-1/2}GD^{-1/2}} \in \mathbb{S}_+.
\end{equation}
By the design of our pseudo signed graph, the corresponding Laplacian will pulls the samples from the same class ($S_{ij} = 1$) closer while pushing others embedding ($S_{ij} = -1$) away towards the opposite direction of each other.
}

\
\textbf{Pseudo cross-view Laplacian with contrast.}
However, learned from clustering without any label supervision, the quality of the pseudo graph highly depends on the quality of feature and can be vulnerable to class-irrelevant noise or any other real-world distortions. 

\textbf{(1) Data augmentation}: To enhance the robustness of pseudo labeling, we first introduce class-preserve augmentations that augment data samples in ways that maintain their original class labels, such as random masking and adding noise. It can effectively smooth out class-irrelevant noise from samples \citep{rebuffi2021data}. As a result, we get the augmented $\widetilde{\mathcal{X}}$ from the original $\mathcal{X}$.

\textbf{(2) Cross-view Laplacian}: Suppose $\mathcal{X}$ and $\widetilde{\mathcal{X}}$ are of the same sample size $N$ corresponding to the input tensor and the augmented tensor, with feature matrices $\mathbf{W} \in \mathbb{R}^{N \times R}$ and $\widetilde{\mathbf{W}} \in \mathbb{R}^{N \times R}$, respectively. \re{To transform the traditional same-view Laplacian penalty in Eq. (\ref{eq:laplacian}) into our desired cross-view, since our augmentation doesn't change the class belonging, which means $\mathbf{w}^{(n)}$ and $\widetilde{\mathbf{w}}^{(n)}$ admit the same class: then,} we can simply replace all $\langle \frac{\mathbf{w}^{(i)}}{\left\|\mathbf{w}^{(i)}\right\|_{2}}, \frac{\mathbf{w}^{(j)}}{\left\|\mathbf{w}^{(j)}\right\|_{2}} \rangle$ \re{(both from the original data)}
with $\langle \frac{\mathbf{w}^{(i)}}{\left\|\mathbf{w}^{(i)}\right\|_{2}}, \frac{\widetilde{\mathbf{w}}^{(j)}}{\left\|\widetilde{\mathbf{w}}^{(j)}\right\|_{2}} \rangle$ \re{(the former $\mathbf{w}$ from original view, the latter $\widetilde{\mathbf{w}}$ from augmented view)} and reformulate the Laplacian loss in Eq. \eqref{eq:laplacian} as:

\begin{equation}  \label{eq:cross_laplacian}
    \ell_{\textit{cross-view}} = \operatorname{Tr(\mathbf{W}^T\mathbf{\Lambda}(\mathbf{W})\textcolor{orange}{\mathbf{L}}\mathbf{\Lambda}(\widetilde{\mathbf{W}})\widetilde{\mathbf{W}}}).
\end{equation}

\begin{wrapfigure}{1}{0.4\textwidth}
    \centering
    \vspace{-10pt}
    \includegraphics[width=0.38\textwidth]{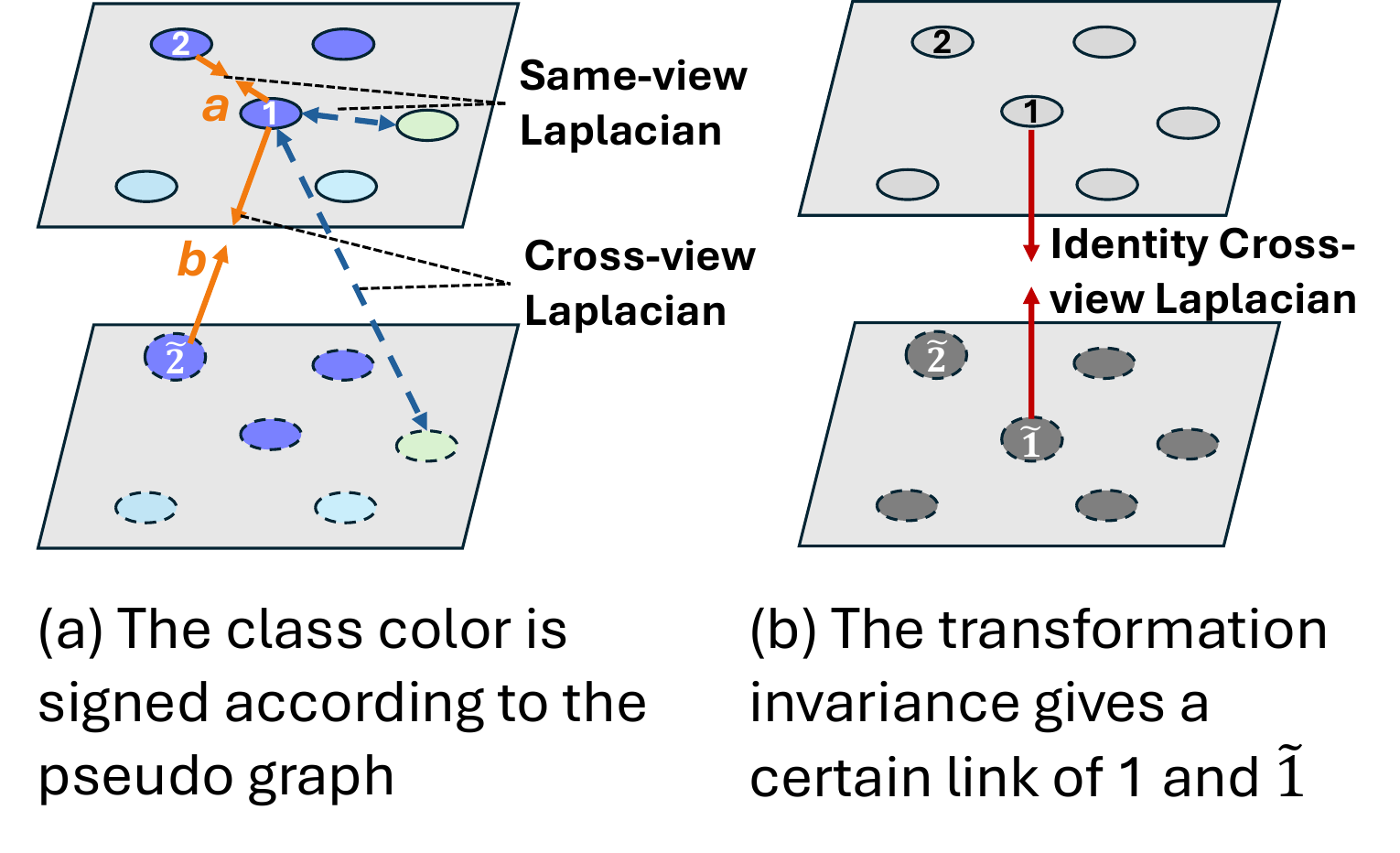}
    \vspace{-8pt}
    \caption{The design of PLC loss.}
    \label{fig:cross-within}
    \vspace{-12pt}
\end{wrapfigure}

\textbf{(3) Transformation invariance is an identity cross-view Laplacian}: As shown in Fig \ref{fig:cross-within}(b), the transformation invariance \re{means the certain alignment between the original sample, e.g., 1, and the augmented sample, e.g., $\Tilde{1}$, even without the knowledge of class or pseudo-class}. It is essential to include an additional loss to ensure that the features from the corresponding raw data and augmented data are also close to each other. It is worth mentioning that transformation invariance is equivalent to cross-view Laplacian penalty with a special Laplacian matrix = $\mathbf{I}$. Thus, we have:
\begin{equation}  \label{eq:trans_inv}
    \ell_{\textit{trans-inv}} = \frac{1}{N}\operatorname{Tr(\mathbf{W}^T\mathbf{\Lambda}(\mathbf{W}) \textcolor{red}{\mathbf{I}} \mathbf{\Lambda}(\widetilde{\mathbf{W}})\widetilde{\mathbf{W}}}) 
\end{equation}
%where $\gamma$ controls the impact. %Actually, the transformation invariance loss is also a special case of the cross-view graph Laplacian with a different graph.

Thus, our final pseudo cross-view Laplacian penalty can be written as:
\begin{equation}\label{eq:l_PLC}
    \ell_{\textit{PLC}} =  -  \ell_{\textit{trans-inv}} + \gamma\ell_{\textit{cross-view}}
\end{equation}

\textbf{(4) ``Cross-view Laplacian = same-view Laplacian + robustness''}: \re{Here we explain why only using cross-view Laplacian penalty is enough, rather than the traditional same-view  Laplacian. 

Firstly, they have the equivalent effect on promoting coherence within the class. As shown in Fig \ref{fig:cross-within}, edge \textit{b} from the cross-view has exactly the same effect as edge \textit{a} from the same-view: both pushing two features from the same class (sample 1 and 2/$\Tilde{2}$) together, }
%We adopt the cross-view graph Laplacian loss $\ell_{PLC}$ rather than the traditional intra-view graph Laplacian due to: \man{we may need a graph to show the relationship between the cross-view edge and intra-view edge; explain why we use cross-view rather than intra-view here more specifically}
\re{ 
Considering that the distance between two features from the same view can be bounded by:
\begin{equation} 
    \| w^{(i)} - w^{(j)}\|_F^2 \leq \| w^{(i)} - \Tilde w^{(i)}\|_F^2 +\| w^{(j)} - \Tilde w^{(i)}\|_F^2.
\end{equation}
If $S_{ij} = 1$, our cross-view PLC loss tries to minimize both terms on the right which will in turn push the samples from the same class closer. If $S_{ij} = -1$, the first term will be minimized while the second term will be maximized which can pull away the two samples under the adjustment of hyper-parameter $\gamma$.

Furthermore, in edge \textit{b}, noise and other distortions are further incorporate into sample $\Tilde{2}$, which encourage robustness additionally. Thus, not like the over-design of $\ell_{\textit{same-view}}+\ell_{\textit{cross-view}}$, which will introduce twice computational complexity, we only employ the simple yet effective  $\ell_{\textit{cross-view}}$.
}

\textbf{Overall loss function.}
Considering the principles of fitness and alignment, the overall objective integrates the our PLC loss, CP decomposition reconstruction loss, and also parameter regularization:

\begin{equation}  \label{eq:loss}
    L  = \ell_{\textit{cp}} + \alpha \ell_{\textit{reg}} + \beta\ell_{\textit{PLC}},
\end{equation}

where $\ell_{\textit{cp}} =  \frac{1}{2} \| \mathcal{X} - \llbracket \mathbf{W}, \mathbf{A}, \mathbf{B}, \mathbf{C} \rrbracket \|^2 
    + \frac{1}{2} \| \widetilde{\mathcal{X}} - \llbracket \widetilde{\mathbf{W}}, \mathbf{A}, \mathbf{B}, \mathbf{C} \rrbracket \|^2$, the regularization term $\ell_{\textit{reg}} =  \frac{1}{2}(\|\mathbf{\mathbf{W}}\|_{F}^{2} +\|\mathbf{\widetilde{\mathbf{W}}}\|_{F}^{2} + \|\mathbf{\mathbf{A}}\|_{F}^{2} + \|\mathbf{\mathbf{B}}\|_{F}^{2} + \|\mathbf{\mathbf{C}}\|_{F}^{2})$ to reduce the model complexity and avoid overfitting \citep{g:97}, $\alpha, \beta, \gamma\geq0$ are hyperparameters that can be determined via the AIC/BIC criterion.

\begin{comment}
\begin{equation*}\label{eq:loss}
\begin{aligned}
    \ell_{cp}& =  \frac{1}{2} \| \mathcal{X} - \llbracket \mathbf{W}, \mathbf{A}, \mathbf{B}, \mathbf{C} \rrbracket \|^2 
    + \frac{1}{2} \| \widetilde{\mathcal{X}} - \llbracket \widetilde{\mathbf{W}}, \mathbf{A}, \mathbf{B}, \mathbf{C} \rrbracket \|^2,\\
    \ell_{reg}& =  \frac{1}{2}(\|\mathbf{\mathbf{W}}\|_{F}^{2} +\|\mathbf{\widetilde{\mathbf{W}}}\|_{F}^{2} + \|\mathbf{\mathbf{A}}\|_{F}^{2} + \|\mathbf{\mathbf{B}}\|_{F}^{2} + \|\mathbf{\mathbf{C}}\|_{F}^{2}), \\
    \Tilde{\ell}_{ss}& = \operatorname{Tr(\mathbf{W}^T\mathbf{\Lambda}(\mathbf{W})\mathbf{L}\mathbf{\Lambda}(\widetilde{\mathbf{W}})\widetilde{\mathbf{W}}})
    - \gamma\operatorname{Tr(\mathbf{W}^T\mathbf{\Lambda}(\mathbf{W})\mathbf{\Lambda}(\widetilde{\mathbf{W}})\widetilde{\mathbf{W}}}),
\end{aligned}
\end{equation*}
\end{comment}

%where $\alpha, \beta, \gamma\geq0$ are hyperparameters that can be determined via the AIC/BIC criterion, the regularization loss trying to reduce the model complexity and avoid overfitting \citep{g:97}.

Intuitively, the loss function jointly optimizes three objectives. The first two terms are trying to restoring the original tensor from the shared low-rank bases. The third term is trying to push feature vectors from the same class closer to the each other from both cluster-wise and instance-wise view.

\subsection{Deep Dive: Relationship with contrastive learning}

\re{
\textbf{The proposed method is equivalent to a constrained contrastive learning method.} Suppose the feature vectors $\mathbf{w}$ are normalized, we compare the pseudo-Laplacian loss, $\ell_{\textit{PLC}}$, with the popular InfoNCE contrastive loss \cite{oord2018representation}:
\begin{equation}  \label{eq:infonce}
    \ell_{\textit{InfoNCE}} = -\frac{1}{N} \sum_{i = 1} ^ {N} \mathbf{w}^{(i)^T}\Tilde{\mathbf{w}}^{(i)} + \frac{\tau}{N} \sum_{i=1}^{N} \operatorname{log} \sum_{i=1}^N \operatorname{exp}(\mathbf{w}^{(i)^T}\Tilde{\mathbf{w}}^{(i)} /\tau).
\end{equation}

The proposed cross-view Laplacian penalty can be expanded into squared form by:
\begin{equation}  \label{eq:sqaure}
    \ell_{\textit{PLC}} = -\frac{1}{N} \sum_{i = 1} ^ {N} \mathbf{w}^{(i)^T}\Tilde{\mathbf{w}}^{(i)} + \frac{\gamma}{N-1} \sum_{i=1}^{N} \sum_{i=1}^N (-S_{ij}) \cdot \mathbf{w}^{(i)^T}\Tilde{\mathbf{w}}^{(i)}.
\end{equation}

It's obvious that both loss function pull the positive pairs and push negative pairs towards the same and opposite direction of each other for contrast, but with different definition of the positive sampling strategies. 
InfoNCE only considers the pair-wise contrastive loss that only $\mathcal{X}$ and its direct augmented sample $\widetilde{\mathcal{X}}$ are denoted as positive sample, while the pseudo graph Laplacian involve the latent class information by letting all the samples from the same distribution to be positive pairs. 
This makes PLC more powerful to obtain class-distinctive features. 
}

\re{
\textbf{Block-contrastive Loss.} By leveraging groups of positive data for each sample according to the pseudo labeling, it further enjoys the property of block-contrastive loss which is always bounded by the pairwise-contrastive under the same loss definition.
% The proposed PLC loss leverages blocks of similar data for each sample according to the pseudo labeling, rather than only instance pairs, thereby it further enjoys the property of block-contrastive loss which is always bounded by the pairwise-contrastive \cite{arora2019theoretical}. 
Given the feature $\mathbf{w^{(n)}}$ of a specific sample, the inequality always holds:
% (will be discussed in Appendix \ref{App:tech}):
\begin{equation}  \label{eq:block}
    -\mathbf{w}^{(i)^T} (\sum_{S_{ij}=1} S_{ij}\Tilde{\mathbf{w}}^{(j)} - \sum_{S_{ij}=-1}S_{ij}\Tilde{\mathbf{w}}^{(j)}) \leq -\mathbf{w}^{(i)^T} (S_{ii}\Tilde{\mathbf{w}}^{(i)} - \sum_{i,j}|S_{ij}|\Tilde{\mathbf{w}}^{(j)}).
    % \frac{\sum_i \widetilde{\mathbf{w}}^{(n)}_{i+}}{c} - \frac{\sum_i \widetilde{\mathbf{w}}^{(n)}_{i-}}{c'}) = \sum_n -\mathbf{w}^{(n)^T}  (\mathbf{w}^{(n)}_+ - \mathbf{w}^{(n)}_-)
\end{equation}
$\mathbf{w}^{(i)}_+ = \sum_{S_{ij}=1} S_{ij}\Tilde{\mathbf{w}}^{(j)} $ and $ \mathbf{w}^{(i)}_- = \sum_{S_{ij}=-1}S_{ij}\Tilde{\mathbf{w}}^{(j)}$ are denoted as the positive and negative blocks respectively. Intuitively, for each sample, the PLC loss will loop over all the positive samples from the same class and negative samples from different classes defined by learned pseudo graph $\mathcal{G}$.
% $\widetilde{\mathbf{w}}^{(n)}_{i+}$ and $\widetilde{\mathbf{w}}^{(n)}_{i-}$ from its positive samples and negative samples accordingly, Eq. \ref{eq:sqaure} can be transformed to (refer to Suppl. Material for details):
% \begin{equation}
%     \sum_n -\mathbf{w}^{(n)^T} (\frac{\sum_i \widetilde{\mathbf{w}}^{(n)}_{i+}}{c} - \frac{\sum_i \widetilde{\mathbf{w}}^{(n)}_{i-}}{c'}) = \sum_n -\mathbf{w}^{(n)^T}  (\mathbf{w}^{(n)}_+ - \mathbf{w}^{(n)}_-)
% \end{equation}
% where $\mathbf{w}_+ $ and $ \mathbf{w}_-$ represent positive and negative blocks respectively, $c$ and $c'$ are constants for normalizing the block size. Intuitively, for each sample, the PLC loss will loop over all the positive samples from the same class and negative samples from different class defined by the the learned pseudo graph $\mathcal{G}$.
}

\subsection{Unsupervised alternative optimization} \label{sec:opt}
We apply the alternative least squares (ALS) algorithm \citep{s:17} to update mode matrices $\mathbf{A}$, $\mathbf{B}$,$\mathbf{C}$, features  $\mathbf{W}$, $\widetilde{\mathbf{W}}$  iteratively, and then use clustering method for pseudo labeling in an unsupervised manner. 

\textbf{Closed-form for $\mathbf{W}$ and $\widetilde{\mathbf{W}}$ with linear convergence.} 
Given $\mathbf{A}, \mathbf{B}, \mathbf{C}$ fixed and pseudo Laplacian matrix $\mathbf{L}$, the feature vector $\mathbf{w}^{(n)}$ can be solved via the following optimization problem:
% $\mathbf{w}^{(n)}$ can be solved via the following optimization problem:
\begin{equation} 
\underset{\mathbf{w}^{(n)}}{\operatorname{argmin}}\left( \alpha\|\mathbf{w}^{(n)}\|_{2}^{2} \right. + \beta \operatorname{Tr}\left( \frac{\mathbf{w}^{(n)^{T}}}{\left\|\mathbf{w}^{(n)}\right\|_{2}} \mathbf{\ell}^{(n)} \mathbf{\Lambda}(\widetilde{\mathbf{W}}) \widetilde{\mathbf{W}} \right)
\left. + \left\| \mathcal{X}^{(n)} - \llbracket \mathbf{w}^{(n)}, \mathbf{A}, \mathbf{B}, \mathbf{C} \rrbracket \right\|_{2}^{2} \right).
\label{w_update}
\end{equation}

\noindent where $\mathbf{w}^{(n)}, \mathbf{\ell}^{(n)}$ are the $n$-th row of $\mathbf{W}, \mathbf{L}$, and $\mathcal{X}^{(n)} $ is the $n$-th slice of $ \mathcal{X}$.
By setting the derivate of Eq. \eqref{w_update} to be zero and arrange the terms, we can get the following updating rule:
% (the derivation can be found in Appendix \ref{App:tech}:
\begin{equation} 
%\begin{aligned}
\mathbf{w}^{(n)} = \left(\mathbf{X}_{(1)}^{(n)} \mathbf{H}_{1}\right.
\left.- \frac{\beta}{2} \mathbf{H}_{2} /\left\|\mathbf{w}^{(n)}\right\|_{2}\left(1-\mathbf{w}^{(n)} \mathbf{w}^{(n)^{T}} /\left\|\mathbf{w}^{(n)}\right\|_{2}^{2}\right)\right) 
\times \left(\mathbf{H}_{1}^{T} \mathbf{H}_{1}+\alpha \mathbf{I}_{R}\right)^{-1},
%\end{aligned}
\label{w_fix}
\end{equation}

\noindent where $\mathbf{H}_{1}=\mathbf{A} \odot \mathbf{B} \odot \mathbf{C} \in \mathbb{R}^{(I J K) \times R}$; $\mathbf{H}_{2}=$ $\mathbf{\ell}^{(n)} \mathbf{\Lambda}(\widetilde{\mathbf{W}}) \widetilde{\mathbf{W}} \in \mathbb{R}^{1 \times R}$; $\mathbf{I}_{R}\in \mathbb{R}^{R \times R}$ is an identity matrix. The fixed point of Eq. \eqref{w_fix} is a stationary point of Eq. \eqref{w_update}. $\mathbf{w}_{n}$ can be updated iteratively until convergence with an initial guess by setting $\beta=0$ in Eq. \eqref{w_update}.
% By subscribing the initial guess of Eq. \eqref{w_update} when $\beta=0$, $\mathbf{w}^{(n)}$ can be updated iteratively until convergence.
% the fixed point is found. 
We can alternatively update  $\widetilde{\mathbf{w}}^{(n)}$ similarly. 
With a good initial $\mathbf{w}^{(n)}_{0}$ (i.e., estimated from only the reconstruction loss), Eq. \ref{w_fix} converges linearly as stated by Proposition \ref{prop:1}.

\begin{prop}{\textbf{Linear Convergence} of $\mathbf{w}^{(n)}$}\label{prop:1}
\textit{{ Given non-zero vectors $\mathbf{H}_{1}$, $\mathbf{H}_{2}$, $\mathbf{w}^{(n)}_{0}$, and $\alpha, \beta >0$, we have
    $
{\left\|\mathbf{w}^{(n)}_{t+1}-\mathbf{w}^{(n)^*}\right\|_{2}}\leq
\frac{\beta (2m+M){\left\|\mathbf{H}_{2}\right\|_{2}}{\left\|\mathbf{H}_{1}^{T} \mathbf{H}_{1}+\alpha \mathbf{I}_{R}\right\|_{F}}}{m^3}{\left\|\mathbf{w}^{(n)}_{t}-\mathbf{w}^{(n)^*}\right\|_{2}}
$, where $t$ represents the $t$-th iteration, $\mathbf{w}^{(n)^*}$ is the fixed point, $m=\min_t{{\left\|\mathbf{w}^{(n)}_{t}\right\|_{2}}}$, $M=\min_t{{\left\|\mathbf{w}^{(n)}_{t}\right\|_{2}}}$ are the bound of the sequence. }}
\end{prop}

% The optimization process can be summarized in an Expectation Maximization (EM) framework - alternatively estimate the pseudo labels and minimize the loss function.

\begin{prop}{\textbf{PCL searches an angle for class preservation within an $\epsilon$-Region of reconstruction loss}}: \textit{To better understand the optimization process, a relaxed optimization problem is:}

\begin{equation} 
    \begin{aligned}
        min  \ \ \ \ &\operatorname{Tr(\mathbf{W}^T\mathbf{\Lambda}(\mathbf{W})\mathbf{L}\mathbf{\Lambda}(\widetilde{\mathbf{W}})\widetilde{\mathbf{W}}})
    - \gamma\operatorname{Tr(\mathbf{W}^T\mathbf{\Lambda}(\mathbf{W})\mathbf{\Lambda}(\widetilde{\mathbf{W}})\widetilde{\mathbf{W}}})\\
        s.t.   \ \ \ \ & \| \mathcal{X} - \llbracket \mathbf{W}, \mathbf{A}, \mathbf{B}, \mathbf{C} \rrbracket \|^2 \leq \epsilon , \ \ \ \ \ \   
         \| \widetilde{\mathcal{X}} - \llbracket \widetilde{\mathbf{W}}, \mathbf{A}, \mathbf{B}, \mathbf{C} \rrbracket \|^2 \leq \epsilon
    \end{aligned}
\end{equation}

\textit{It indicates that given the factorization matrices, we search the best direction in $\mathcal{E}$, which preserves most class information corresponding to the pseudo label. The reconstruction constraints guarantee the quality and consistency of our representations.}
\end{prop}

\begin{wrapfigure}{l}{0.7\textwidth}
    \centering
    \vspace{-20pt}
    \includegraphics[width=0.7\textwidth]{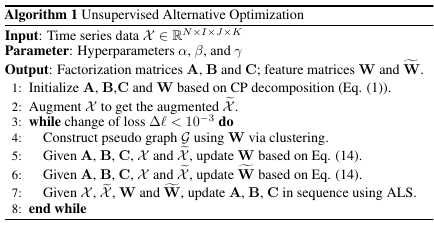}
    \vspace{-15pt}
    %\caption{The design of PLC loss.}
    %\label{fig:alg1}
    \vspace{-12pt}
\end{wrapfigure}
\textbf{Optimization procedures.}
The ALS and pseudo learning are conducted cyclically in a similar Expectation Maximization (EM) framework - alternatively estimate the pseudo labels and minimize the loss function - as summarized in Algorithm 1. 
We alternatively update $\mathbf{A}$, $\mathbf{B}$, $\mathbf{C}, \mathbf{W}$ and $\widetilde{\mathbf{W}}$ using ALS by solving least square subproblems with closed solutions.
The updated feature matrix $\mathbf{W}$ is then used for pseudo labeling via clustering. The overall complexity of each iteration is around $O(N^3R)$, where R is the tensor rank and \re{$N$ is the number of samples}.

\section{Experiments}
\label{sec_experiment}
%The proposed PCT method has been evaluated on downstream classification task to prove the effectiveness and and illustrate its insights.
% under unsupervised setting - without access to label information except for the classifier. The experiment results have well proved our model's effectiveness and insights.

\subsection{Experimental setup} 
\label{sec:exper_setup}
\paragraph{\textbf{Datasets.}}
% \subsection{\textbf{Dataset}}
We use three real-world datasets for validation: 
(i) \textbf{HAR data} is human activity recognition data \citep{a:13} collected as 3-axial linear acceleration and 3-axial angular velocity at a constant rate of $50 \mathrm{~Hz}$ 
% by the embedded accelerometer and gyroscope, with license included in their citation; 
(ii) \textbf{Sleep-EDF data} \citep{k:00} collected 153 whole-night PolySomnoGraphic sleep recordings, containing EEG (from Fpz-Cz and Pz-Oz electrode locations), EOG (horizontal), and submental chin EMG recordings; 
%under Open Data Commons Attribution License v1.0. 
(iii) \textbf{PTB-XL data} \cite{wagner2020ptb} collected 21873 clinical 12-lead ECGs.

\textbf{\textbf{Data preprocessing.}}
\label{para_2_C}
 The datasets have been randomly partitioned into unlabeled, training, and test sets by subjects, where the latter two sets for classifier have labels, with a ratio of 70\%:15\%:15\%. Specifically, we down-sample the dataset to smaller scales. Then Short-Time Fourier Transforms (STFT) is performed after the data augmentation to encode the time series data into three-order tensors of size \textit{channel} $\times$ \textit{frequency} $\times$ \textit{timestamp}, i.e., $18 \times 33 \times 33$ for HAR. 
% Please note that the same data augmentation and preprocessing steps are applied to all methods.

\textbf{Class-preserved augmentations.} As shown in Fig \ref{fig:overview}(a), following \citep{h:20, c:20, d:19}, we apply three different augmentation methods in sequence \re{to get the augmented tensor $\widetilde{\mathcal{X}}$ from the original tensor $\mathcal{X}$}: (i) \textit{Jittering} adds additional perturbations to each sample. 
(ii) \textit{Bandpass} filtering reduces signal noise. We use the order-1 Bandpass filter 
to preserve only the within-band frequency information. 
(iii) \textit{3D position rotation} uses a 3D $x$-$y$-$z$ coordinate system rotation by a rotation matrix to mimic different cellphone positions. 
% And the perturbed data share the same class information.

\textbf{\textbf{Baselines.}}
We group benchmarks into tensor-based and self-supervised models, and apply them in the same framework to validate the effectiveness of the proposed method. Tensor-based methods include CPD \cite{k:09}, ATD \cite{y:22}, and DeepTensor \cite{s:22}, which are our main competitors. Note that ATD is the first method to incorporate self-supervised learning into the tensor decomposition but only considers instance-level alignment, and DeepTensor uses generative networks to enable more intricate tensor decomposition. We also compare two popular self-supervised learning methods, SimCLR \citep{c:20} and BYOL \citep{g:20}, with their own objective functions under the same CNN backbone as \cite{cheng2020subject}.

\textbf{Experiment settings.} 
Under unsupervised setting, all models have no access to label information to train feature extractor except for the downstream logistic classifier \cite{h:20}. 
We use the official implementation for all benchmark methods to train for all the datasets. 
All models are optimized for 100 epochs by Adam optimizer with a learning rate of 0.001. For deep models, the dimension of the hidden state is 128.
For tensor-based models, the rank is R = 32.
% More configurations (e.g., hyper-parameters selection) can be found in Appendix \ref{app: exp}.

\subsection{\textbf{Result analysis}}

\begin{table*}
\centering
\caption{Comparing different methods in downstream classification, with each sample's dimension specified after the dataset. Our model is good at extracting class-aware representations with average \textbf{+5.35\%} higher accuracy over tensor models and more efficient running time per epoch. When handling complex data (Sleep-EDF, PTB-XL), we achieve similar running time as SSL methods.}
% \label{tab:class_acc}
\resizebox{0.95\textwidth}{!}{%
\begin{tabular}{c|c|c|c|c|c|c|c} 
% \hhline{=====}
%  Dataset & HAR  & Sleep-EDF & PTB-XL \\
 \hhline{========}
 \multirow{2}{*}{} &  \multirow{2}{*}{Dataset} & \multicolumn{2}{c|}{\textbf{HAR} ($18\times 33\times 33$) } & \multicolumn{2}{c|}{\textbf{Sleep-EDF} ($14\times 129\times 86$)} & \multicolumn{2}{c}{\textbf{PTB-XL} (($24\times 129\times 75$))} \\
 \cline{3-8}
 & & Accuracy & Time/epoch(s)  & Accuracy & Time/epoch & Accuracy & Time/epoch(s)  \\ 
% \cline{1-11}

\hhline{========}
 % \cline{1-5}
% \hline{===========}
% \textbf{Method} & \multicolumn{6}{c}{\textbf{Classification Accuracy}} & &aware & anomaly & projection \\ 
% \hhline{=======~===}
\multirow{2}{*}{\textbf{SSL}} & SimCLR  
&0.6662$_{3.8e-2}$  & 7.9508
&\underline{0.8313}$_{7.2e-3}$ & 15.7657
& 0.6442$_{6.7e-2}$ & 19.5069

\\
& BYOL 
&0.7017$_{1.0e-1}$ &7.4217
& 0.8257$_{1.2e-1}$ & 15.4009
& 0.6251$_{1.4e-2}$ & 19.260
\\\cline{1-8}

\multirow{4}{*}{\textbf{Tensor}} & CPD 
& 0.8813$_{1.6e-3}$ & 3.6819
& 0.7913$_{3.2e-2}$ & 6.8075
& 0.6749$_{5.7e-2}$ & 7.2089
 \\
& Deep Tensor 
&0.8371$_{5.4e-2}$ & 17.8731
&0.7667$_{1.5e-2}$ & 20.1100
& \underline{0.6932}$_{1.9e-2}$  & 20.6826
\\
& ATD 
&\underline{0.8951}$_{2.6e-2}$  & 16.0915
& 0.8293$_{1.6e-1}$ & 18.2486
& 0.6868$_{2.0e-2}$  &28.7403
\\%\cline{2-8}
 &  \textbf{PLC (Ours)} 
& \textbf{0.9349}$_{6.5e-2}$ & 13.7321
& \textbf{0.8420}$_{1.7e-3}$ & 18.6653
& \textbf{0.7147}$_{7.9 e-2}$  & 20.1542
 \\
\hhline{========}
\end{tabular}%
}
% \caption{Comparison of different methods' classification robustness towards different levels of data corruption: RSDT though is not the strongest method when data is purely clean (which is rather rare in real industry case), RSDT demonstrates the strongest robustness, e.g., with only \textbf{-1.289\%} drop in HAR (10\%, $\times 5$), compared with ATD's \textbf{-15.71\%} drop. Anomaly values are shown as the multiple of signal average.}
% \vspace{-0.3cm}
\label{tab:class_acc}
\end{table*}

We reports the average classification accuracy with standard deviations under four-time repeated experiment and also the average running time per epoch. Several ablation study and visualization have been implemented for illustration.

\begin{figure}[t]
\centering
\vspace{-0.4cm}
\includegraphics[width=\columnwidth]{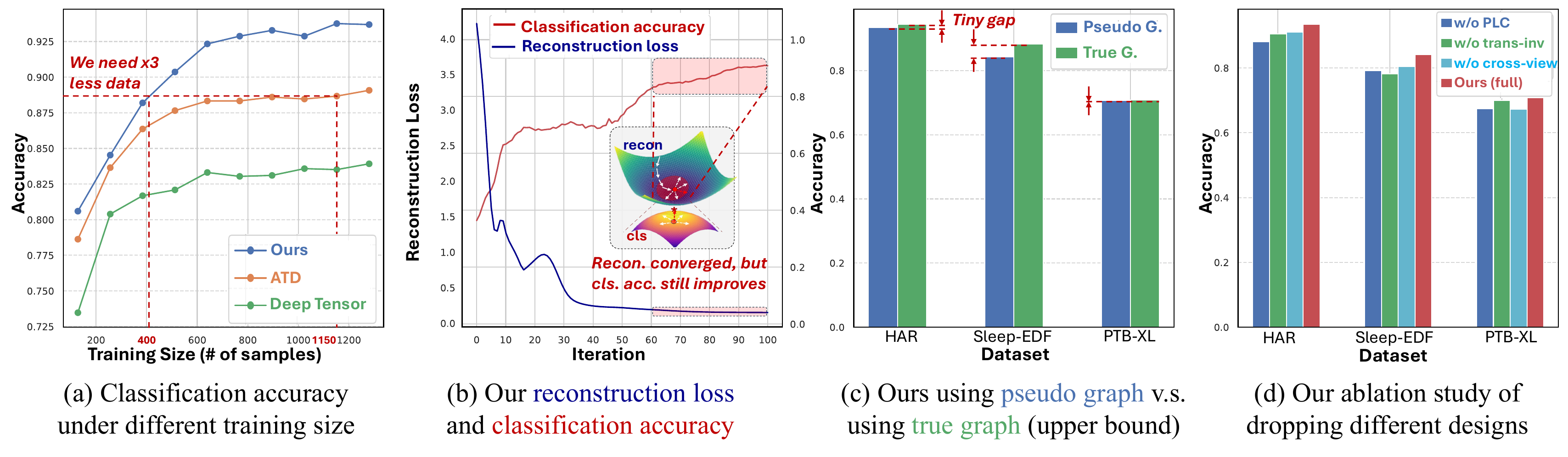}
% \captionsetup{font=small}
\vspace{-15pt}
\caption{ (a) The classification performance over varying training size for classifier: \re{to achieve same accuracy, ours can need 3 times less data than ATD}; (b) The curve of reconstruction loss and classification loss: the two term improve each other and classification accuracy keeps increasing after the convergence of reconstruction loss; (c) Supervised contrast with true label \re{can be seen as the upper bound of our pseudo-graph, and our pseudo-Laplacian contrast has only a very tiny gap with the upper bound}. (d) Ablation study of ``w/o PLC" (no $\ell_{\textit{PLC}}$), ``w/o cross-view" (no $\ell_{\textit{cross-view}}$) and ``w/o trans-inv" (no $\ell_{\textit{trans-inv}}$).}
% \ziyue{(1) the fontsize in the three figures should be larger, (2) the caption should be self-contained.}}
\vspace{-0.5cm}
\label{fig:insight}
\end{figure}

\begin{figure}[t]
  \centering
  \vspace{-0.5cm}
  \begin{minipage}[t]{0.65\textwidth}
    \centering
    \includegraphics[width=\textwidth]{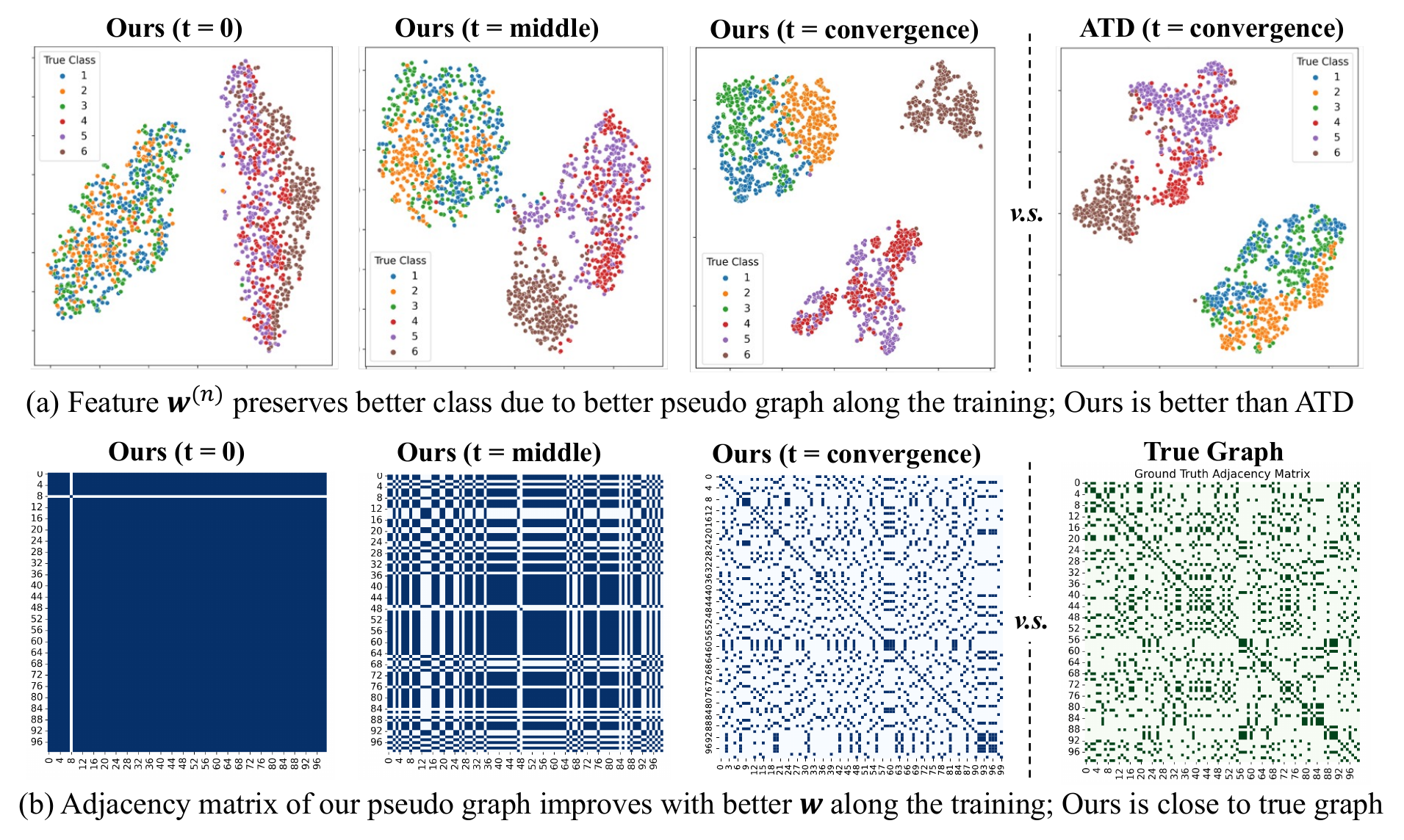}
    % \captionsetup{font=small}
    \vspace{-12pt}
    \caption{The visualization of $\mathbf{w}^{(n)}$ and pseudo graph $\mathcal{G}$ on HAR dataset: (a) the t-SNE of the learned feature at the beginning and convergence where our model learns more clearly-separated class embedding; (b) the adjacent matrix of the pseudo-graph is progressively approaching that of the true graph. }
    % \ziyue{the fontsize inside of the figure should be larger}}
    \label{fig:vis}
  \end{minipage}%
  \hfill
  \begin{minipage}[t]{0.32\textwidth}
    \centering
    \includegraphics[width=\textwidth]{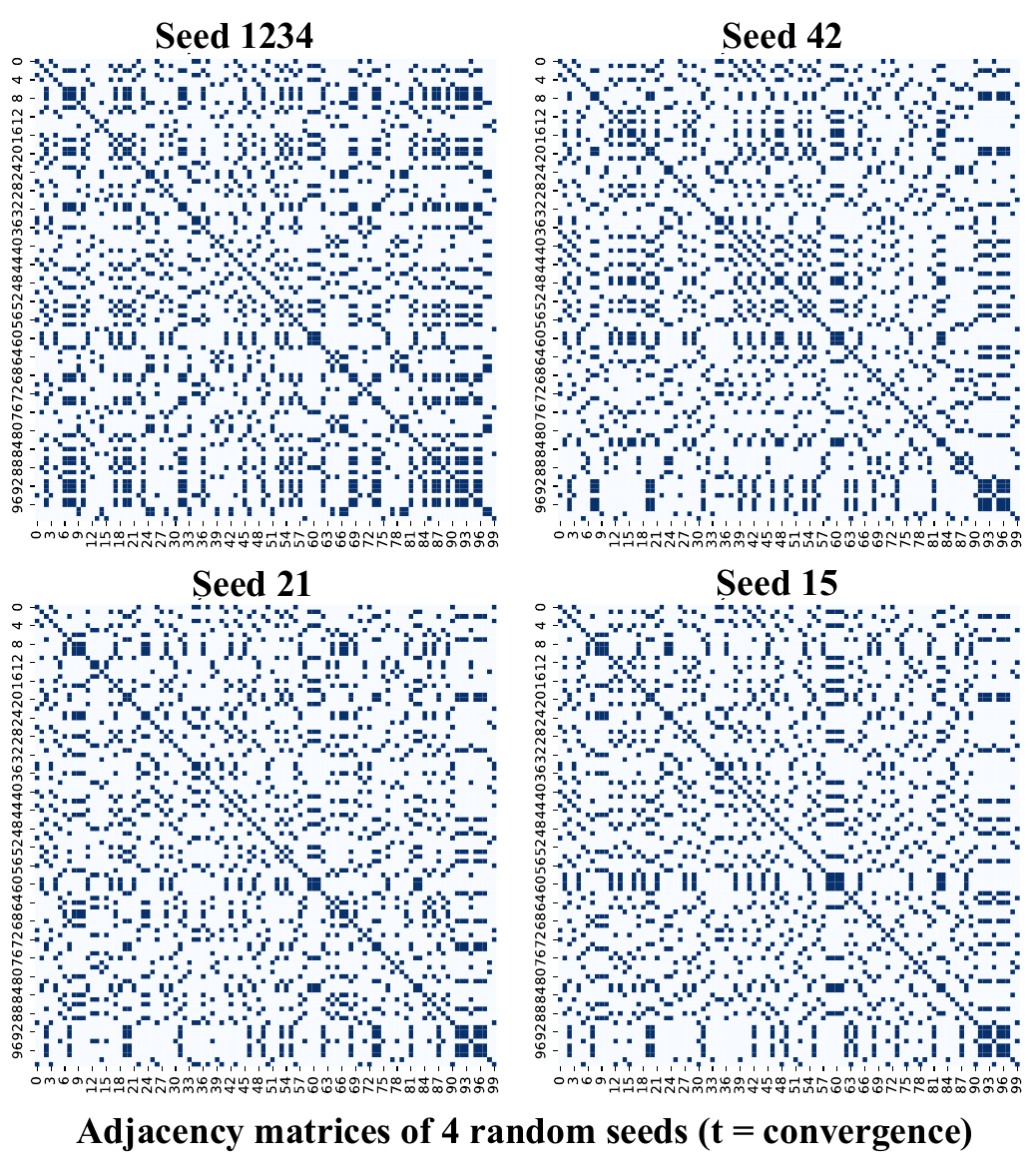}
    % \captionsetup{font=small}

    \caption{The learned pseudo graph under different random seeds of decomposition remains consistent with each other and close to the ground truth.}
    \label{fig:adj}
  \end{minipage}
  \vspace{-0.54cm}
\end{figure}

\textbf{Pseudo-graph guided contrast further boosts the performance.} From Table \ref{tab:class_acc}, we can make several observations for the benefit of contrastive loss: (1) compared with traditional tensor models only supervised by reconstruction loss, contrastive tensor methods generally achieve better performance, e.g.,  ATD improves the accuracy by around +1\% from CPD; (2) our model further achieves \textbf{+4.57\%} average gains over CPD and also \textbf{+1.27$\sim$3.98\%} gains over ATD, which effectively prove that pseudo-graph Laplacian contrastive loss can benefit class-aware feature extraction more.

\textbf{Advantage from data compression ability when handling complex data.}
Table \ref{tab:class_acc} reports the running time per epoch and we can observe that simple CPD is the fastest one due to its dimensionality reduction objective. The two self-supervised learning runs faster than ours, especially when handling HAR data with small sample shape, but the difference shrinkage for more complex data. This presents the advantage of data compression ability to handle complex data, which will be more useful in the real world applications.

% \textbf{Performance Comparison}: 
% From Table \ref{tab:class_acc}, we can make several observations:
% \begin{itemize}
% \setlength\itemsep{0.1em}
%     \item \textbf{Pseudo-graph guided contrast further boosts the performance}: compared with traditional tensor models, e.g., CPD and DeepTensor, ATD and our PCL, with self-supervised loss, generally achieve better classification accuracy. Also, our model can capture class-aware features better and outperforms ATD with \textbf{+1.53$\sim$4.45\%} gains, which effectively prove the benefit of pseudo-graph guided contrastive loss.

%     \item \textbf{Tensor-based models show greater generalization}: the deep self-supervised models is more sensitive to the data size, with relatively poor performance especially on HAR dataset with average \textbf{-36.75\%} worse than ours . The reasons could be (i) their complex deep structure cannot be well optimized on small dataset while the reconstruction loss of tensor decomposition help preserve the intrinsic; (ii) HAR and PTB-XL data may fit the low-rank assumption, enabling better representation. On Sleep-EDF dataset, SimCLR with marginally better accuracy shows that its embedding space remains more information.

% \end{itemize}

\textbf{Better generalization on small training size.} (1) Overall, deep self-supervised models show poorer performance than tensor-based models, especially on HAR dataset with an average \textbf{-25.10\%} negative gain compared to ours. It may be caused by the insufficient samples to well optimize their complex deep models, while the reconstruction loss of tensor decomposition help preserve the intrinsic data characteristics. (2) Figure \ref{fig:insight}(a) can further illustrate the better generalization of our model within tensor-based models, with consistently higher accuracy over varying classifier training size. With access to few labeled data, the performance largely depends on the class-distinctiveness of the embedding space, which shows that our learned representation preserves more class information.
% \ref{fig:class}

\textbf{Reconstruction and contrastive loss are both critical.} As shown in Fig. \ref{fig:insight}(b), there are two stages where the reconstruction loss decreases dramatically and classification acc. increases slowly first; when the reconstruction loss nearly converges, the classification accuracy then keeps increasing further. It proves the deep insights claimed in Figure \ref{fig:overview}(b) that our model is trying to find the subspace that best preserves the class information within the small region with a small reconstruction loss.

\textbf{Comparable performance with the inclusion of true label.} To prove the power of pseudo graph, we compare our performance with a supervised learning paradigm with ground true labels to guide representation learning. As shown in Fig. \ref{fig:insight}(c), there only exists a very tiny drop of the performance on average, which is confident to show the powerful guidance of our pseudo graph, which is leaned close to the ground true one.

\textbf{Transformation invariance and cross-view Laplacian both benefits the accuracy.} Fig. \ref{fig:insight}(d) shows the ablation study that investigates the individual contributions of different components. The ablation study shows that the full PCL loss contributes most to the classification performance by over $\textbf{+3\%}$ improvements on all three datasets. 
Notably, without $\ell_{\textit{trans-inv}}$ in Sleep-EDF, the performance drop dramatically since pseudo-graph labeling may learn the wrong latent class without $\ell_{\textit{trans-inv}}$ to enforce robustness. 
The contribution of $\ell_{\textit{trans-inv}}$ and $\ell_{\textit{cross-view}}$ have different degrees over three datasets.

\textbf{The feature space is class-distinctive.} Fig. \ref{fig:vis} presents the visualization of feature vectors $\mathbf{W}$ and pseudo graph $\mathcal{G}$ on HAR data. We could observe that (1) the representation learned by our model is more class-distinctive with separated cluster patterns from the t-SNE visualization; (2) the adjacent matrix of the pseudo graph is progressively approaching the true graph which indicates that the optimization process keeps trying to learn the correct pseudo labels and successfully approximate the ground true graph in an unsupervised manner. This again explains why our model have comparable performance with the true-graph guided paradigm.

\begin{comment}
\begin{wrapfigure}{1}{0.4\textwidth}
    \centering
    \vspace{-10pt}
    \includegraphics[width=0.4\columnwidth]{figure/ablation_pcl.pdf}
% \captionsetup{font=small}
\caption{Ablation study: (a) `w/o PLC", ``w/o cross-view" and ``w/o trans-inv" are denoted as removing our $\ell_{\textit{PLC}}$, $\ell_{\textit{cross-view}}$ and $\ell_{\textit{trans-inv}}$ respectively; (b) the choice of hyperparameter $\gamma$ have great impact on the performance.}
    \label{fig:ablation}
    \vspace{-10pt}
\end{wrapfigure}

\begin{figure}[t]
\centering
% \vspace{-0.5cm}
\includegraphics[width=0.8\columnwidth]{figure/ablation_pcl.pdf}
% \captionsetup{font=small}
\caption{Ablation study: (a) `w/o PLC", ``w/o cross-view" and ``w/o trans-inv" are denoted as removing our $\ell_{\textit{PLC}}$, $\ell_{\textit{cross-view}}$ and $\ell_{\textit{trans-inv}}$ respectively; (b) the choice of hyperparameter $\gamma$ have great impact on the performance.}
% \ziyue{(1) the fontsize in the three figures should be larger, (2) the caption should be self-contained.}}
\vspace{-0.5cm}
\label{fig:ablation}
\end{figure}
\end{comment}
\textbf{The learned pseudo graph is consistently close to the true graph.} In Fig. \ref{fig:adj}, we visualize the adjacent matrix from the learned pseudo graph under different random seeds of decomposition and observe that they are almost the same and close to the ground truth shown in Fig. \ref{fig:vis}. This is a strong evidence of our model's strong capability for class information extraction.

\section{Conclusion} \label{sec:con}
The proposed Pseudo Laplacian Contrast (PLC) Tensor Decomposition presents a simple but powerful framework for time-series classification. We design a cross-view Laplacian loss based on the learned pseudo graph and theoretically validate that it is equivalent to contrastive learning. Through iteratively improving pseudo graph and latent feature, our model can effectively and consistently learn a class-distinctive embedding space with pseudo graph close to the true graph.
Extensive experimental results on various datasets demonstrate that PLC achieves competitive performance over a wide range of tensor-based and self-supervised models regarding both classification accuracy and generalization.
Our work shows the great potentials of tensor decomposition for classification, however it may not work well when class information is not consistent with the data correlation itself.
More in-depth explorations will be conducted in future work.
% ?We introduce graph wiener filter and theoretically validate its superior reconstruction ability to facilitate reconstruction-based representation learning.
% By leveraging graph wiener decoder, our model can efficiently learn graph embedding with augmentation. 
% Extensive experimental results on various datasets demonstrate that  WGDN achieves competitive performance over a wide range of self-supervised and semi-supervised counterparts. 

\newpage
\bibliographystyle{plain}
\bibliography{main}

\end{document}